\documentclass[review]{elsarticle}
\usepackage{lineno}
\usepackage{graphicx}
\usepackage{amssymb}
\modulolinenumbers[5]

\journal{Neurocomputing}

\graphicspath{{epsFig/}}

\begin{document}

\begin{frontmatter}

\title{Feature Learning from Incomplete EEG with Denoising Autoencoder}


\author[mymainaddress]{Junhua Li \corref{mycorrespondingauthor}}
\cortext[mycorrespondingauthor]{Corresponding Author}
\ead{juhalee.bcmi@gmail.com}
\author[mymainaddress]{Zbigniew Struzik}
\author[mysecondaryaddress]{Liqing Zhang}
\author[mymainaddress]{Andrzej Cichocki}


\address[mymainaddress]{Laboratory for Advanced Brain Signal Processing, Brain Science Institute, RIKEN, Saitama, 351-0198, Japan}
\address[mysecondaryaddress]{Key Laboratory of Shanghai Education Commission for Intelligent Interaction and Cognitive Engineering, Department of Computer Science and Engineering, Shanghai Jiao Tong University, Shanghai, 200240, China}

\begin{abstract}
An alternative pathway for the human brain to communicate with the outside world is by means of a brain computer interface (BCI). A BCI can decode electroencephalogram (EEG) signals of brain activities, and then send a command or an intent to an external interactive device, such as a wheelchair. The effectiveness of the BCI depends on the performance in decoding the EEG. Usually, the EEG is contaminated by different kinds of artefacts (e.g., electromyogram (EMG), background activity), which leads to a low decoding performance. A number of filtering methods can be utilized to remove or weaken the effects of artefacts, but they generally fail when the EEG contains extreme artefacts. In such cases, the most common approach is to discard the whole data segment containing extreme artefacts. This causes the fatal drawback that the BCI cannot output decoding results during that time. In order to solve this problem, we employ the Lomb-Scargle periodogram to estimate the spectral power from incomplete EEG (after removing only parts contaminated by artefacts), and Denoising Autoencoder (DAE) for learning. The proposed method is evaluated with motor imagery EEG data. The results show that our method can successfully decode incomplete EEG to good effect.        
\end{abstract}


\end{frontmatter}

\linenumbers

\section{Introduction}

\noindent The combination of advanced neurobiology and engineering creates a new pathway, namely a brain computer interface (BCI). The BCI provides a bridge connecting the human brain to the outside world \cite{ortiz2013brain}. This means that people do not have to rely on the conventional pathway of an intent initialized in the brain being passed to muscles through peripheral nerves, and are able to interact directly with the external environment \cite{wolpaw2000brain}. Due to the lack of involvement of peripheral nerves and muscles, with the aid of a BCI system, disabled people could restore their abilities of communication \cite{muller2008machine} and the degenerated motor function \cite{li2013design, pfurtscheller2003thought}. During the past two decades, a variety of BCI systems have been created for different applications. These BCI systems are generally divided into two types: active BCI and passive BCI, according to the level of interaction with external stimuli. In the case of a passive BCI, when using a steady-state visual evoked potential (SSVEP) BCI \cite{bin2009online}, the user may, for example, simply stare at an intended digital number shown on a screen to dial a phone number. When a steady-state flicker is replaced with an occasional flicker, a different type of BCI called P300 can be used to output letters by hierarchical selections \cite{muller2008machine}. Compared to the passive BCI, the active BCI is more natural. Users can express their intents whenever they want to, rather than according to a predefined timing arrangement or external cooperation, as with the passive BCI. For instance, people with paraplegia can regain movement in a wheelchair by motor imagery \cite{li2013design}, or can control a computer cursor in virtual 2D \cite{mcfarland1993eeg} or 3D \cite{mcfarland2010electroencephalographic} environments through brain modulation. Moreover, BCI is also used to develop prostheses, with which disabled people can, for example, move an object \cite{muller2005eeg} or drink a cup of coffee \cite{hochberg2012reach}. More recently, BCI has been applied to facilitate rehabilitation \cite{daly2008brain, liu2014tensor}. Besides applications for disabled people, BCI also has promising applications for healthy persons, especially in the field of entertainment. BCI is employed to control video games instead of conventional inputs such as a keyboard and joystick \cite{li2013competitive}. In this way, healthy people can enjoy the experience of manipulating virtual objects in a manner different from that used in daily life.          
\\
\noindent From the application point of view, the user experience is very important. This requires smoothness in the manipulation of the BCI system. In order to meet this requirement, the BCI system needs to translate brain activities into output information continuously without any interruption. In other words, this requires all the EEG segments to be present for the decoding. If some of the EEG segments are discarded due to extreme noise contamination, the BCI cannot generate the corresponding output during that period. Hence, it would be good to be able to utilize the remaining portion of the affected EEG segment, after only removing the part directly affected by noise. 
In general, spectral power features are usually utilized to distinguish different motor imageries (e.g., left-hand and right-hand motor imageries) \cite{palaniappan2006utilizing, pfurtscheller2000brain, li2010bilateral, li2012active}, as they are considered to be robust for the representation of the contents of motor imageries. If the segment is complete (continuous), the Fourier transform can be well used to transform temporal data points into the spectral domain. This fails in the case of incomplete data, such as an EEG segment with a portion (or portions) of data removed (unevenly spaced). In order still to utilize such segments of EEG with arbitrary portions of data removed and provide users with an experience of smooth manipulation, we employ the Lomb-Scargle periodogram to estimate the spectral power \cite{lomb1976least, stoica2009spectral}, and Denoising Autoencoder (DAE) \cite{vincent2008extracting,vincent2010stacked} based neural network or support vector machine (SVM) \cite{vapnik2000nature, hearst1998support} to predict the classes of motor imageries. The results show that the proposed method is suitable for decoding incomplete EEG in a BCI system.

\section{Methodology}
\begin{figure*}[!htb]
\centering
\includegraphics[width=1\textwidth]{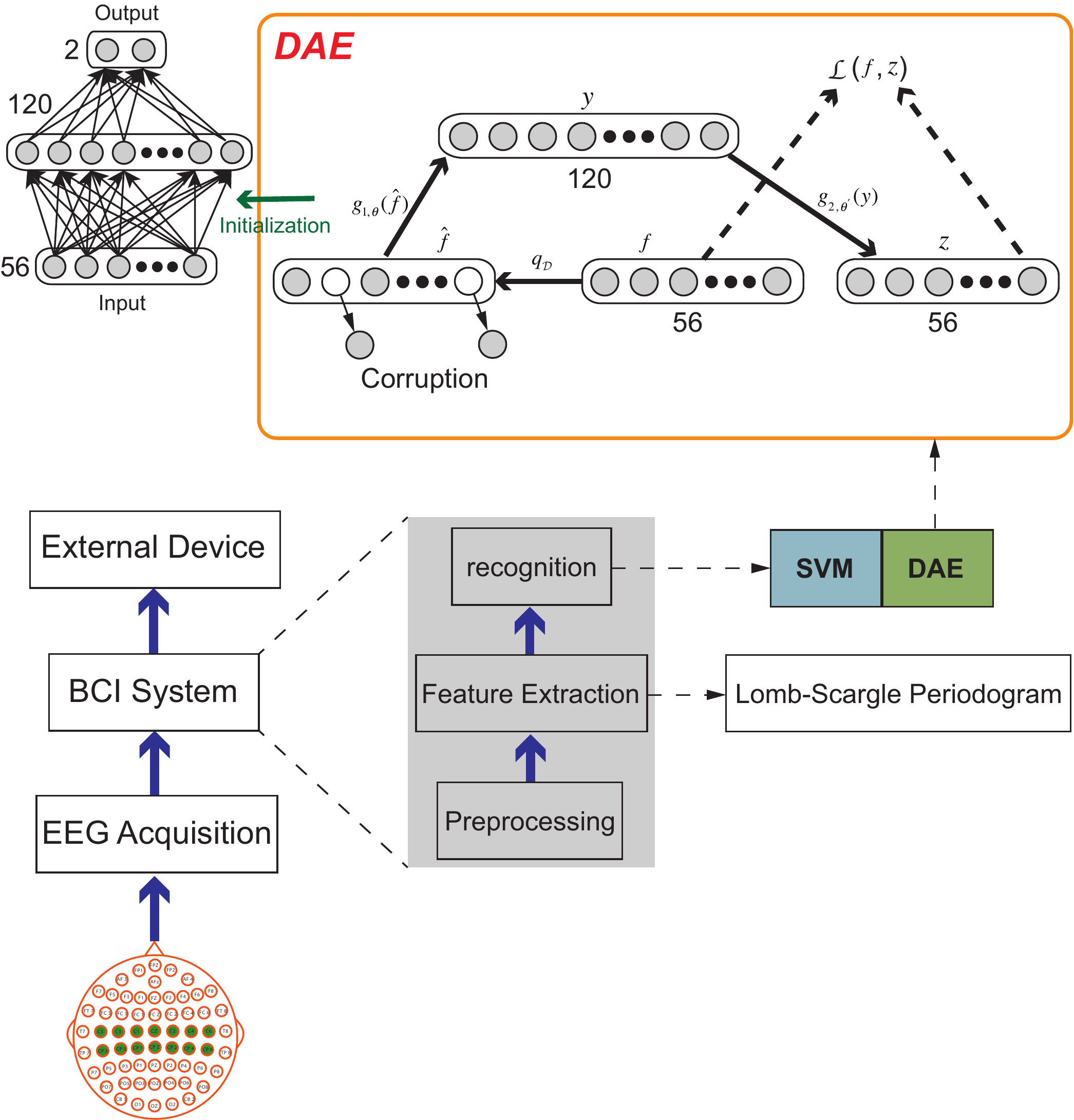}
\caption{\label{schematic} Schematic depiction of the proposed method.}
\end{figure*}

\noindent We first employed the Lomb-Scargle periodogram \cite{lomb1976least, stoica2009spectral} to estimate band powers from incomplete EEG signals. Next, the extracted features were used to train an unsupervised DAE  \cite{vincent2008extracting,vincent2010stacked} or a supervised SVM with Radial Basis Function (RBF) kernel \cite{vapnik2000nature, hearst1998support}. In the case of DAE, the mapping weights of DAE were used to initialize a neural network. After fine-tuning the weights, this trained neural network was used to recognize the classes of motor imageries. Fig. \ref{schematic} illustrates the proposed method.

\subsection{Lomb-Scargle Periodogram}
\noindent A four-second trial is divided into 25 segments of one-second length with an overlap of 87.5\%. A segment is denoted by $X$, which is $N$ by $T$  matrix, where $N$ is the number of channels, and $T$ is the number of sampling points. The spectral power of each channel time series $y({t_i})$ is estimated by the Lomb-Scargle periodogram \cite{lomb1976least, stoica2009spectral}. The estimated spectral power at frequency ${\Omega_f}$  can be obtained by minimizing the following sum of difference squares:

\begin{equation}
\label{equation1}
\mathop {\min }\limits_{\scriptstyle\;\;\;\;a > 0\hfill\atop
\scriptstyle\phi \in [0,\;2\pi ]\hfill} \;\sum\limits_{i = 1}^T {{{(y({t_i}) - \alpha \cos ({\Omega_f}{t_i} + \phi ))}^2}}\;.
\end{equation}

\noindent Let
\begin{equation}
a = \alpha \cos \phi 
\end{equation}
and
\begin{equation}
b =  - \alpha \sin \phi \;.
\end{equation}

\noindent We can then rewrite equation (\ref{equation1}) as:
\begin{equation}
\label{equation4}
\mathop {\min }\limits_{a,\;b} \;\sum\limits_{i = 1}^T {{{(y({t_i}) - a\cos ({\Omega_f}{t_i}) - b\sin ({\Omega_f}{t_i}))}^2}} .
\end{equation}
The optimal parameters $\hat{a}$ and $\hat{b}$ can be obtained through minimizing equation (\ref{equation4})
\begin{equation}
\left[ {\begin{array}{*{20}{c}}
{\hat{a}}\\
{\hat{b}}
\end{array}} \right] = {R^{ - 1}}r,
\end{equation}
where
\begin{equation}
R = \sum\limits_{i = 1}^T {\left[ {\begin{array}{*{20}{c}}
{\cos ({\Omega_f}{t_i})}\\
{\sin ({\Omega_f}{t_i})}
\end{array}} \right]} \left[ {\begin{array}{*{20}{c}}
{\cos ({\Omega_f}{t_i})}&{\sin ({\Omega_f}{t_i})}
\end{array}} \right],
\end{equation}
and
\begin{equation}
r = \sum\limits_{i = 1}^T {\left[ {\begin{array}{*{20}{c}}
{\cos ({\Omega_f}{t_i})}\\
{\sin ({\Omega_f}{t_i})}
\end{array}} \right]} \;y({t_i})\;.
\end{equation}
The power of specific frequency ${\Omega_f}$ is then estimated with respect to optimal parameters $\hat{a}$, $\hat{b}$ as follows:
\begin{equation}
\begin{array}{l}
\frac{1}{T}{\sum\limits_{i = 1}^T {\left( {\left[ {\begin{array}{*{20}{c}}
{\hat{a}}&{\hat{b}}
\end{array}} \right]\left[ {\begin{array}{*{20}{c}}
{\cos (\Omega{}_f{t_i})}\\
{\sin (\Omega{}_f{t_i})}
\end{array}} \right]} \right)} ^2}\\
  \quad = \frac{1}{T}\left[ {\begin{array}{*{20}{c}}
{\hat{a}}&{\hat{b}}
\end{array}} \right]\;R\;\left[ {\begin{array}{*{20}{c}}
{\hat{a}}\\
{\hat{b}}
\end{array}} \right]\\
 \quad = \frac{1}{T}{r^{\rm T}}({\Omega_f}){R^{ - 1}}({\Omega_f})r({\Omega_f})\;.
\end{array}
\end{equation}

\noindent Similarly, the minimization of squares mentioned above is used to estimate spectral powers at all frequencies. After that, spectral estimation for one channel is completed. These steps are repeated for all channels and all segments to obtain the spectral powers. Because the frequency range of 8-30 Hz is mostly related to the motor imagery task \cite{li2012active}, we divided this band into four subbands with a bandwidth of 5 Hz (i.e., 8-12 Hz, 13-17 Hz, 18-22 Hz, and 23-27 Hz). Subband powers were obtained by averaging spectral powers within the corresponding frequency band range for each channel. Then, subband powers (four features for each channel) for all channels were concatenated into a feature vector:
\begin{equation}
F = {[{f_{11}},\;{f_{12}},\;{f_{13}},\;{f_{14}},\;{f_{21}},\;{f_{22}},\;{f_{23}},\;{f_{24}}, \cdots ,\;{f_{N1}},\;{f_{N2}},\;{f_{N3}},\;{f_{N4}}]^{\rm T}}\;,
\end{equation}
where $N$ is the number of channels. Subsequently, features were normalized as:
\begin{equation}
{f_{qp}} = \log \left( {\frac{{{f_{qp}}}}{{\sum\limits_{i = 1}^N {\sum\limits_{j = 1}^4 {{f_{ij}}} } }}} \right)\;.
\end{equation}
The normalized features were then fed into a neural network with DAE initialization, or into an SVM classifier to distinguish which class the current EEG segment belongs to.

\subsection{DAE-based neural network}
\noindent For a time, neural networks were less frequently used due to the drawback that they easily became stuck in the local minima, so more use was made of SVM classifier. However, recently neural networks have regained popularity, in particular when using a pre-training strategy \cite{vincent2010stacked, erhan2010does, glorot2011domain}. In this paper, we construct a three-layer neural network with DAE initialization (A neural network with more layers might possibly achieve a better performance through in-depth feature learning). 
\\
\noindent The power features extracted by Lomb-Scargle Periodogram was first corrupted, denoted as $\hat{f}$, by means of a stochastic mapping $\hat{f} \thicksim q_\mathrm{\mathcal{{D}}}(\hat{f}|f)$. The part enclosed by the orange rectangle in Fig. \ref{schematic} shows a schematic diagram of the DAE. We set the corrupted elements to 0. Then, the corrupted features were mapped to a hidden representation (120 units) by the sigmoid function
\begin{equation}
y = g_{1,\;\theta }(\hat{f}) = s(W \cdot \hat{f} + b).
\end{equation} 
Consequently, we reconstructed the uncorrupted $z$ as
\begin{equation}
z=g_{2,\;{\theta ^\prime }}(y).
\end{equation}
The objective was to train parameters $\theta = \{ W, b \}$ and $\theta^{\prime} = \{ W^{\prime}, b^{\prime} \}$ for minimization of the average reconstruction error over a training set. In other words, to find the parameters to let $z$ be as close as possible to $f$, we performed the following optimization:
\begin{equation}
\begin{array}{l}
[{\theta^*},{\theta^{{\prime}\;*}}] = \mathop {\arg \min }\limits_{\theta ,\;{\theta^{\prime}}} \frac{1}{n}\sum\limits_{i = 1}^n {L({f^{(i)}},\;{z^{(i)}})} \\
 \quad \quad \quad \quad = \mathop {\arg \min }\limits_{\theta ,\;{\theta^{\prime}}} \frac{1}{n}\sum\limits_{i = 1}^n {L({f^{(i)}},\;{g_{2,\;{\theta^{\prime}}}}({g_{1,\;\theta }}({\hat{f}^{(i)}})))},
\end{array}
\end{equation}
where $L$ is a squared error loss function $L(f,z) = {\left\| {f - z} \right\|^2}$, n is the number of training samples, and ${\theta^*},{\theta^{{\prime}\;*}}$ are the optimal values of ${\theta ^{}},{\theta^{\prime}}$. Once the optimal parameters were obtained, we were able to use those parameters to initialize a neural network. A top layer was added onto the neural network. After that, the parameters were fine-tuned in a supervised way.

\section{Evaluation Data}
\noindent Two different categories of data are used to prove the feasibility of the proposed method. One is the simulated data and the other is the two-class motor imagery data. We use simulated data to illustrate systematically that spectral power can be correctly estimated when the data become unevenly spaced after removing some data points from them. Further, we use real motor imagery data to demonstrate that classification accuracy does not dramatically decrease when increasing the percentage of data within the segment that has been removed, so that the proposed method is useful to process incomplete data in a BCI system. 
\\
\noindent The simulated data were generated by mixing two sinusoidal signals, which were 3 Hz and 6 Hz, respectively. The maximal amplitude of the 3 Hz sinusoidal signal was 1.5 times that of the 6 Hz sinusoidal signal. The motor imagery data came from three subjects. Fourteen electrodes (shown with a green background in the scalp illustration in Fig. \ref{schematic}) were used to record the EEG signal on the sensorimotor cortex while the subject was conducting motor imagery at a sampling rate of 250 Hz. Those electrodes were referenced at the mastoids behind the ears and grounded at AFz. Each subject participated in four sessions. Each session consisted of 15 trials, each of which was four seconds long. The subject conducted either left hand motor imagery or right hand motor imagery according to the cue shown on the computer monitor.

\section{Results}
\subsection{Simulated data}
\begin{figure*}[!htb]
\centering
\includegraphics[width=1\textwidth]{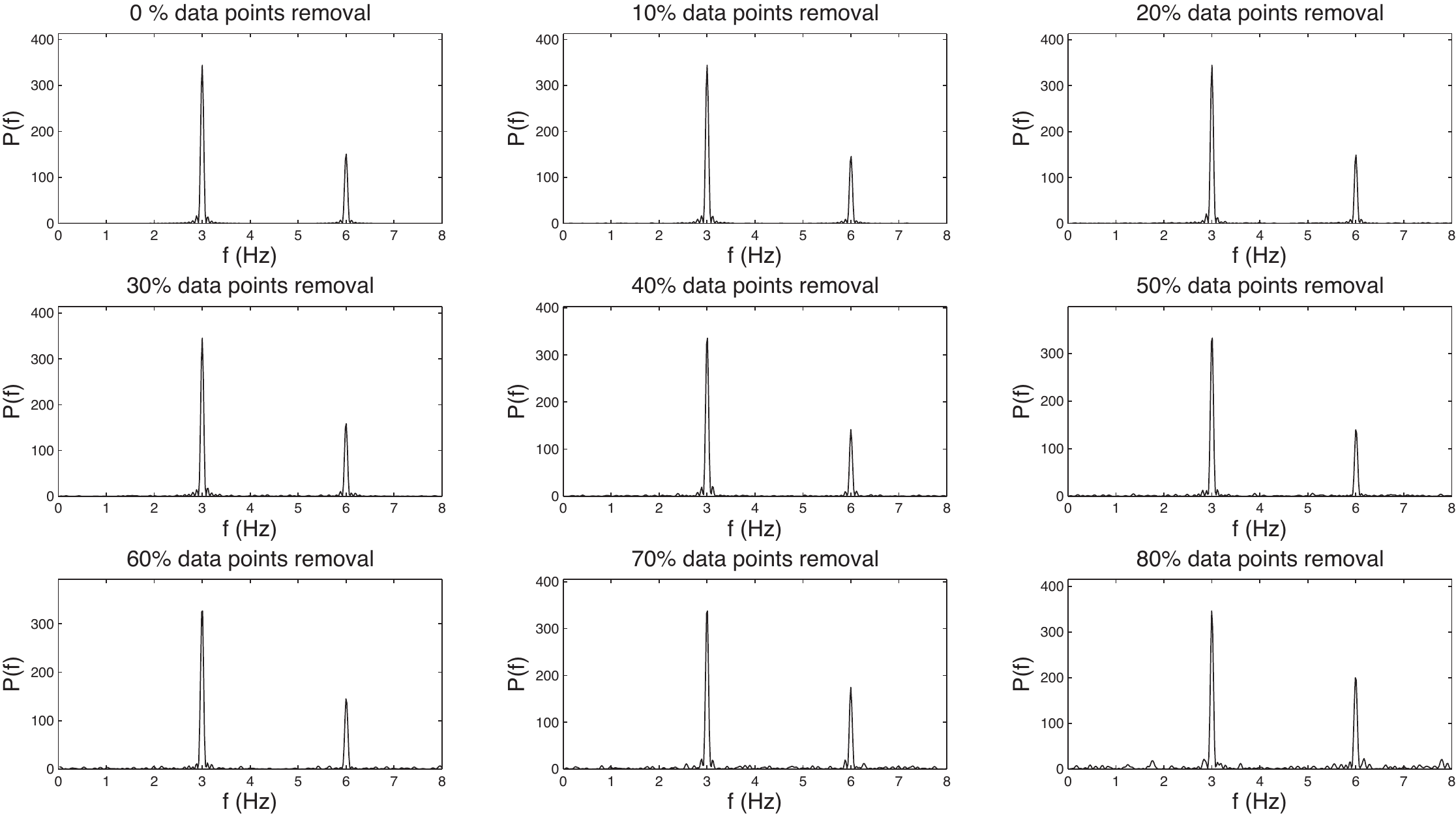}
\caption{\label{SimulatedPlot} Spectral power estimations for the complete signal and signals after data point removal.}
\end{figure*}
\noindent We first evaluated the performance of the spectral power estimation on simulated data. The simulated data was mixed with two sinusoidal signals, which were 3 Hz and 6 Hz, respectively. The spectral power estimated from the complete signal, and the incomplete signals with different proportional removal of data points (from 10\% to 80\% with an interval of 10\%) are shown in Fig. \ref{SimulatedPlot}. The data points were removed at random. In order to keep the same scale over cases with different proportional data removal to facilitate comparisons between them, the powers shown in Fig. \ref{SimulatedPlot} were normalized by dividing by a proportional factor (1-p, where p is the percentage of data removed). For example, the estimated power is divided by a factor of 0.7 when 30\% of data points are removed from the signal. From Fig. \ref{SimulatedPlot}, we can see that the components at 3 Hz and 6 Hz can be well estimated in all cases with different proportions of data removal, even up to removal of 80\% of data points.

\subsection{Real motor imagery data}

\begin{figure*}[!b]
\centering
\includegraphics[width=1\textwidth]{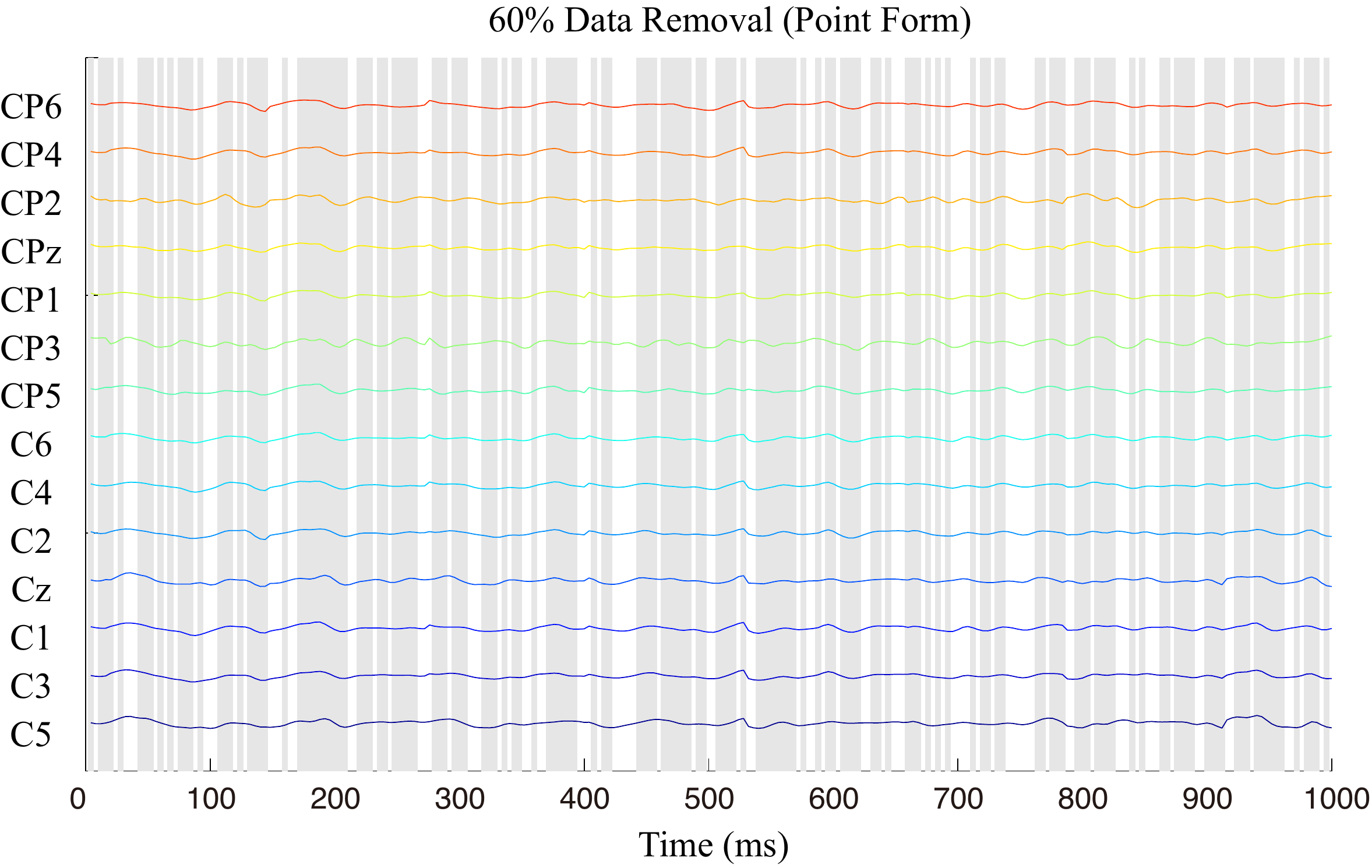}
\caption{\label{Point60Rem} An example of data point removal. The data points shown with a grey background are removed while data points shown with a white background are retained.}
\end{figure*}

\noindent In general, BCI encounters a common problem that there is no output when the whole segment has to be discarded due to partial noise contamination in that segment. If a method can obtain comparable recognition accuracy (the same or slightly worse) by using only the remaining portion of the segment (the portion from which noise contamination has been removed), this method is considered as an effective solution to the aforementioned problem. 

\begin{figure*}[!htb]
\centering
\includegraphics[width=1\textwidth]{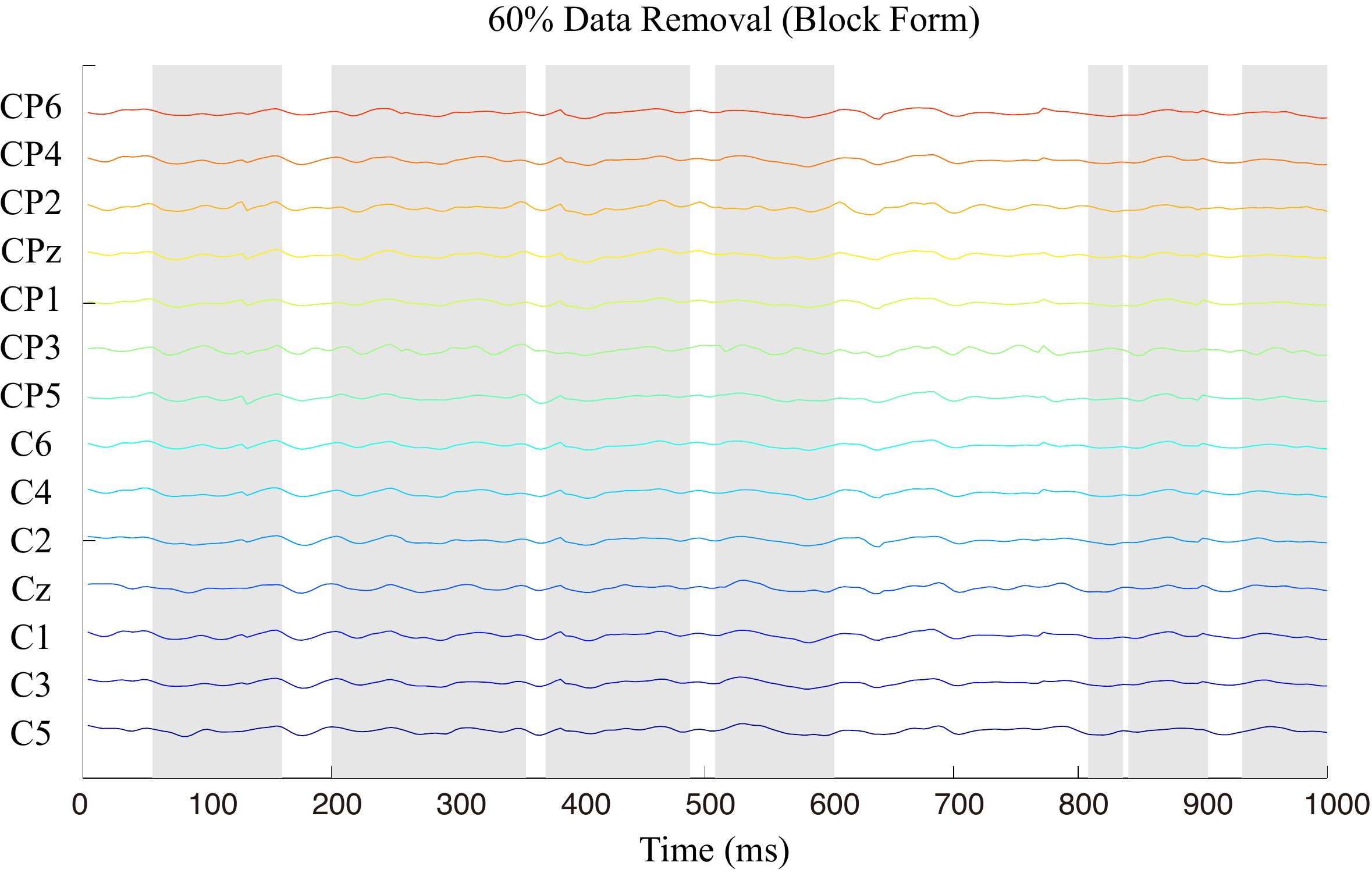}
\caption{\label{Block60Rem} An example of block point removal. The data points shown with a grey background are removed, while data points shown with a white background are retained.}
\end{figure*}

\noindent For real motor imagery data, two ways were used to randomly remove the partial data from the segment. One is that data points within a segment were randomly removed (see Fig. \ref{Point60Rem} for an example). The other is that data blocks within a segment were randomly removed (see Fig. \ref{Block60Rem} for an example). The width of the blocks removed was generated according to a normal distribution with a mean of 20 and a standard deviation of 10.

\begin{figure*}[!thb]
\centering
\includegraphics[width=1\textwidth]{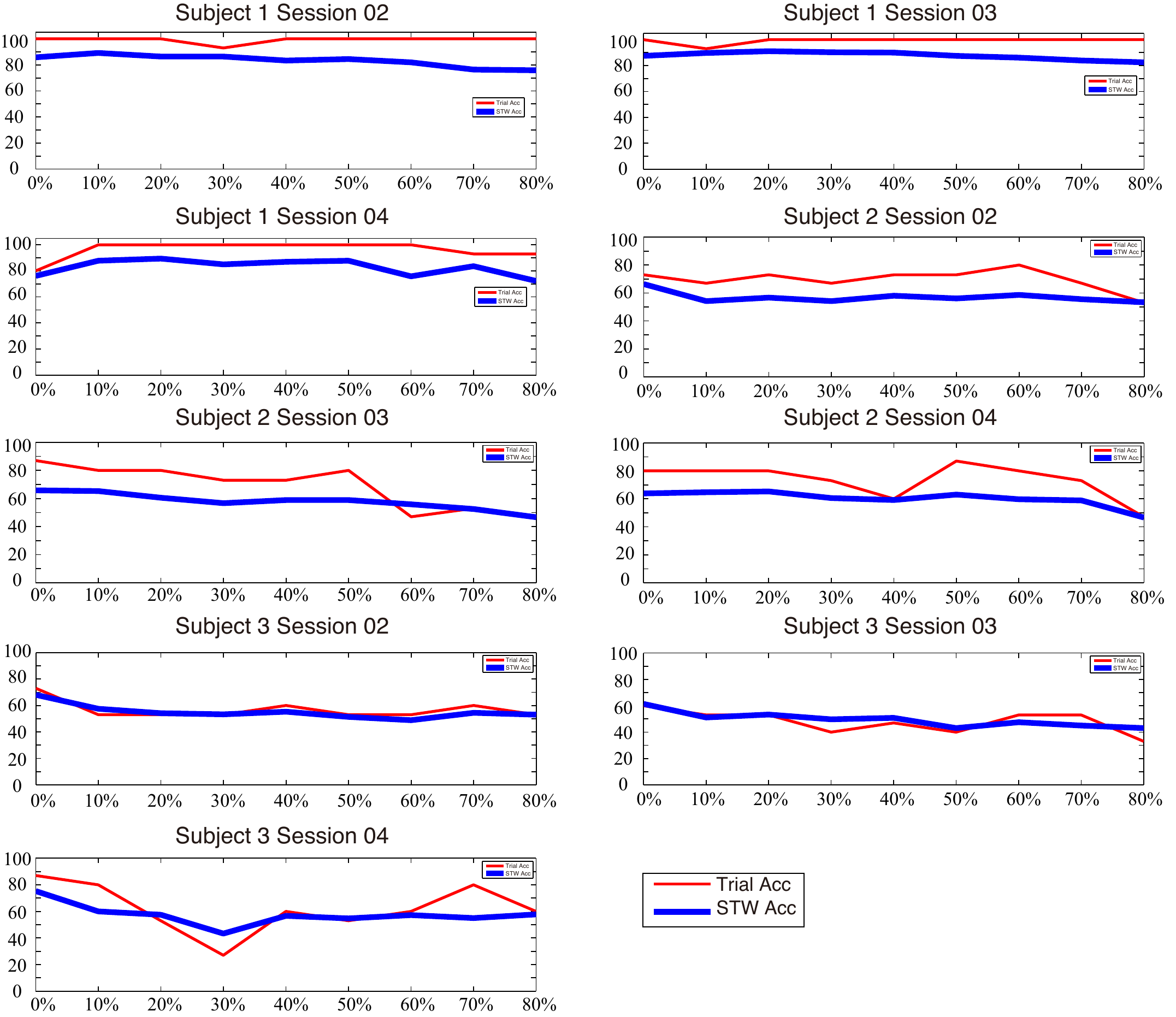}
\caption{\label{AccPoint} Classification accuracies for the form of data point removal. The thin red lines represent trial accuracies, and the bold blue lines represent sliding time window accuracies.}
\end{figure*}

\begin{figure*}[!htb]
\centering
\includegraphics[width=1\textwidth]{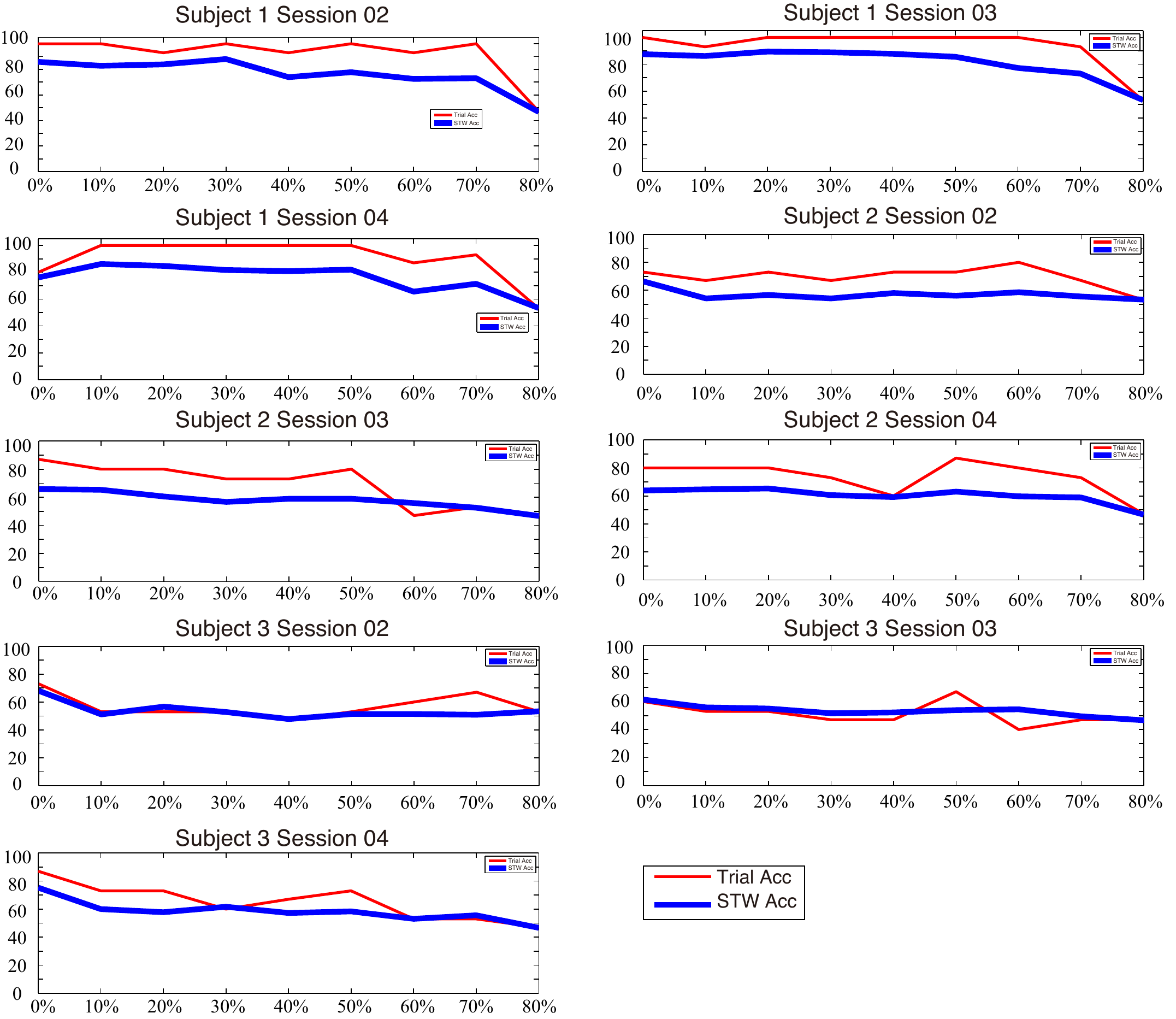}
\caption{\label{AccBlock} Classification accuracies for the form of block point removal. The thin red lines represent trial accuracies, and the bold blue lines represent sliding time window accuracies.}
\end{figure*}

\noindent We used the data from the preceding session to train the SVM classifier with a RBF kernel, and tested it with the data from the following session. Two approaches were used for the evaluation of the accuracy (i.e., sliding time window accuracy and trial accuracy). Sliding time window accuracies were calculated as the number of segments classified as correct divided by the total number of segments. A trial was classified as belonging to the class to which most of the sliding time windows within that trial belonged. Then trial accuracies were obtained according to the ratio of trials classified as correct. Fig. \ref{AccPoint} and Fig. \ref{AccBlock} show test accuracies for the conditions of data point removal and data block removal, respectively. In general, the accuracies for all sessions of all subjects did not dramatically decrease. Trial accuracies varied more than sliding time window accuracies across different proportional sections of data removal. This is because a trial was classified as correct even if the number of sliding time windows in the trial classified as correct was only one more than the number of sliding time windows classified as incorrect. Likewise, trials with one more incorrect sliding time window than correct sliding time window were classified as incorrect. Therefore, in some cases, the trial accuracy changed greatly while the accuracy of 
the sliding time widows did not change much. A comparable classification accuracy could be achieved even when 80\% of data were removed. High accuracies were retained no matter how many data points were removed - in the range from 10\% to 80\% - for subject 1, especially for sessions 2 and 3. The accuracies for 80\% data removal were substantially worse than those for 70\% data removal for subject 1 in the condition of block data removal. It appears that our method is relatively sensitive to data removal in block form.

\subsection{Comparison between DAE and SVM}

\begin{figure*}[!hbt]
\centering
\includegraphics[width=1\textwidth]{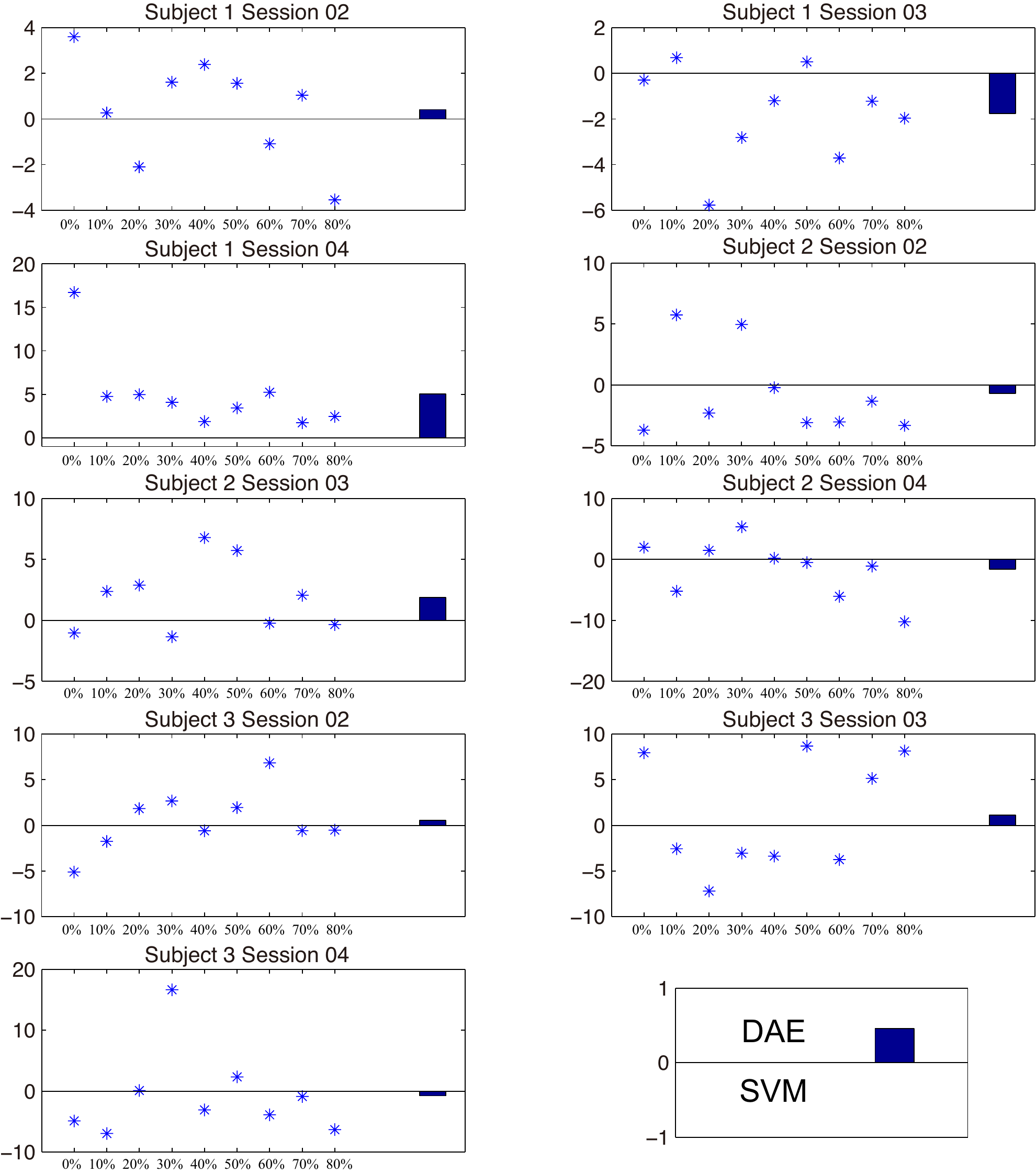}
\caption{\label{AccComPoint} Accuracy comparison between DAE and SVM under the condition of data point removal. Each asterisk represents an accuracy difference between the DAE and the SVM. The difference is calculated by the DAE accuracy minus the corresponding SVM accuracy. The bar at the right of each plot illustrates the average difference in a session.}
\end{figure*}

\begin{figure*}[!htb]
\centering
\includegraphics[width=1\textwidth]{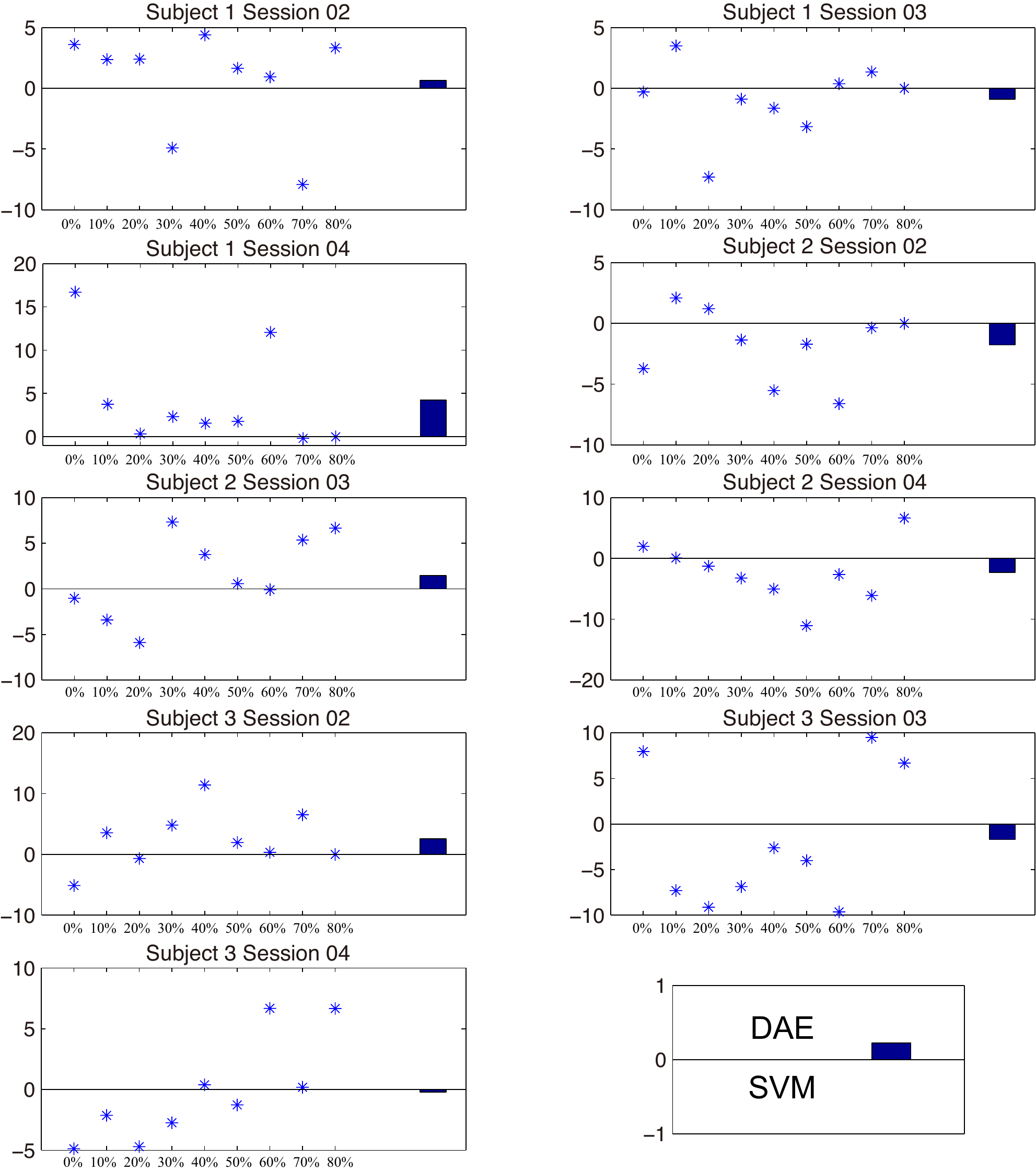}
\caption{\label{AccComBlock} Accuracy comparison between DAE and SVM under the condition of block point removal. Graphical symbol expressions are the same as in Fig. \ref{AccComPoint}.}
\end{figure*}

\noindent In this section, we show a comparison between DAE and SVM in terms of classification accuracy 
of sliding time windows. SVM has been widely adopted since its conception and has been successfully applied in many fields. Deep learning is a promising and burgeoning method. DAE is utilized as a building brick to compose a deep learning network. It is meaningful to illustrate the effectiveness of this for EEG feature recognition using our paradigm. The parameters used in the training are listed in Table \ref{Paras}. Fig. \ref{AccComPoint} shows the accuracy difference between DAE and SVM for each session of each subject under the condition of data point removal. Asterisks located above the zero horizontal line mean that the accuracy of DAE is higher than that of SVM. The bars shown on the right of each sub-plot are the average differences. The bottom right plot illustrates the overall difference averaged across all sessions of all subjects. From Fig. \ref{AccComPoint}, we can see that there is no clear winner - the DAE is better than the SVM in a number of sessions but turns out to be worse in other sessions. The overall average accuracy of DAE is still better than that of SVM. Fig. \ref{AccComBlock} shows the accuracy comparison under the condition of block point removal. The result is similar to the condition of data point removal. The overall average accuracy of DAE is higher than that of SVM under the condition of block point removal, but the increase in accuracy of DAE compared with SVM is less than the case of data point removal. 

\begin{table}[!htbp]
\centering
\caption{Parameter Settings} 
\label{Paras}
\renewcommand\arraystretch{1}
\begin{tabular}{|c|c|}
\hline
Parameters & Values\\ \hline
Corrupted fraction & 0.3\\
Mini-batch size & 25\\
Learning rate for pre-training & 0.9\\
Number of pre-training epochs & 20\\
Learning rate for fine-tuning & 0.9\\
Number of fine-tuning epochs & 50\\ \hline
\end{tabular}
\end{table}

\noindent From the results of comparisons, the DAE is shown to be comparable to the SVM. However, it is possible that the DAE can outperform the SVM when more layers are used and parameters are better tuned. It is not yet clear whether the DAE can significantly exceed the SVM in terms of EEG classification, but there has been a report that stacked DAE (i.e., multiple DAEs combined together to obtain deeper learning of features) performed better than the SVM on the image benchmark dataset named MNIST \cite{vincent2008extracting}.

\section{Conclusion}
\noindent We propose the combination of the Lomb-Scargle periodogram and either SVM or DAE to distinguish incomplete EEG segments (i.e., segments from which a portion of data has been removed due to noise contamination). The results indicate that classification accuracy is not dramatically decreased when different percentages of data are removed. Therefore, the classification performance using the proposed method for incomplete segments is acceptable for a BCI application system. This means that the segment with noise contamination can still be utilized to output commands after only removing the noisy portion, instead of discarding the whole segment, as is conventionally done in BCI systems. In brief, the proposed method can achieve comparable classification performance even when most of the data points in a segment have been removed. It provides an alternative solution for the frequent problem occurring in a BCI system that there is no output when a segment is discarded.

\section{Acknowledgments}
\noindent The work of Liqing Zhang was supported by the national natural science foundation of China (Grant No. 91120305, 61272251).

\section*{References}


\textbf{Junhua Li} received his Ph.D. degree from the Department of Computer Science and Engineering, Shanghai Jiao Tong University, Shanghai, China, in March 2013. He is currently a research scientist at the Laboratory for Advanced Brain Signal Processing, Brain Science Institute, RIKEN, Japan. His research interests include signal processing, brain computer interface, and machine learning. He has been a member of IEEE since 2013, and was a student member of IEEE in 2012.\\

\textbf{Zbigniew R. Struzik} received a Master of Science in Engineering degree
in technical physics from the Warsaw University of Technology, Poland,
in 1986, and a Doctor degree from the faculty of Mathematics, Computer
Science, Physics and Astronomy at the University of Amsterdam, the
Netherlands, in 1996. From 1997 to 2003, he worked at the Centre for
Mathematics and Computer Science (CWI), Amsterdam, and since 2003 has worked at the University of Tokyo, Japan, where he is currently affiliated. From
2012, his main position has been at RIKEN Brain Science Institute in
Wakoshi, Japan. His scientific work contributed to the amalgamation of
(multi-)fractal analysis, wavelet analysis and time series data
mining. His current research interests include applications of
information science and statistical physics in life sciences,
complexity and emergence, time series processing and mining, and
recently, analytic approaches to elucidating the nature of creative
processes in art and science, in particular in neuroscience. He is on
the editorial board of the Fractals Journal, the Open Medical
Informatics Journal, Frontiers in Fractal Physiology, Frontiers in
Computational Physiology and Medicine, Frontiers in Human
Neuroscience, International Journal of Statistical Mechanics, Journal
of Neuroscience Methods and Integrative Medicine International
Journal. He has co-authored over one hundred journal papers and book
chapters.\\

\textbf{Liqing Zhang} received the Ph.D. degree from Zhongshan University, Guangzhou, China, in 1988. He was promoted in 1995 to the position of full professor at South China University of Technology. He worked as a research scientist at RIKEN Brain Science Institute, Japan from 1997 to 2002. He is now a Professor with Department of Computer Science and Engineering, Shanghai Jiao Tong University, Shanghai, China. His current research interests cover computational theory for cortical networks, brain-computer interface, perception and cognition computing model, statistical learning and inference. He has published more than 170 papers in international journals and at conferences.\\

\textbf{Andrzej Cichocki} received Ph.D. and Dr.Sc. (Habilitation) degrees, in electrical engineering, from Warsaw University of Technology (Poland). He is currently the senior team leader head of the laboratory for Advanced Brain Signal Processing, at RIKEN Brain Science Institute (JAPAN). \\
    He is a co-author of more than 250 technical papers and 4 monographs (two of them translated to Chinese). According to a 2011 analysis, he is a co-author of one of the top 1\% most highly cited papers in his field worldwide.

\end{document}